# Spatial Entity Resolution between Restaurant Locations and Transportation Destinations in Southeast Asia

Emily Gao, Dominic Widdows

*Grab Inc.*

Abstract: As a tech company, Grab has expanded from transportation to food delivery, aiming to serve Southeast Asia with hyperlocalized applications. Information about places as transportation destinations can help to improve our knowledge about places as restaurants, so long as the spatial entity resolution problem between these datasets can be solved. In this project, we attempted to recognize identical place entities from databases of Points-of-Interest (POI) and GrabFood restaurants, using their spatial and textual attributes, i.e., latitude, longitude, place name, and street address. Distance metrics were calculated for these attributes and fed to tree-based classifiers. POI-restaurant matching was conducted separately for Singapore, Philippines, Indonesia, and Malaysia. Experimental estimates demonstrate that a matching POI can be found for over 35% of restaurants in these countries. As part of these estimates, test datasets were manually created, and RandomForest, AdaBoost, Gradient Boosting, and XGBoost perform well, with most accuracy, precision, and recall scores close to or higher than 90% for matched vs. unmatched classification. To the authors' knowledge, there are no previous published scientific papers devoted to matching of spatial entities for the Southeast Asia region.

## 1 INTRODUCTION

Location matters in many businesses and services today, particularly for transportation and delivery, scenarios in which it is important to find the correct pick-up and drop-off locations very quickly. User experience can be negatively affected if the location information is inaccurate or insufficient. Inaccuracies can originate from imprecise GPS data, manual error happening in the process of data entry, or the lack of effective data quality control. Insufficiencies can also take many forms, including lack of coverage, and lack of detail — for example, we may know the latitude and longitude of a restaurant location in a mall, but this might not include information about where passengers should be dropped off, or where a delivery courier should park to collect food for delivery. Or the location of a business may be known, but not its contact details or opening hours.

One core problem in managing and improving spatial datasets is recognizing when two records refer to the same real-world entity. Solving this problem can improve precision by removing duplicates, and can enrich detail by (for example) merging a phone number from one record with the hours of operation from another, once these records are known to refer to the same thing. This problem is referred to as entity resolution (see (Talburt, 2011)), and it occurs with various datasets, including those representing people, products, works of literature, etc.

For Grab, one entity resolution problem that arises for spatial data is the alignment of transportation destinations and restaurants. Currently Grab maintains two tables separately for transportation and food delivery, because each use case requires some specific features, i.e., food delivery needs information about the estimated delivery time, cuisine types, and opening hours which are absent in the POI table. However, it is highly likely that some entities from both tables refer to the same place, and how to figure that out is related to the study of entity resolution.

Matching restaurant and POI is beneficial for some use cases in Grab. The first one is about automatic geolocation correction update that could happen in both transportation and food delivery. As shown in Figure 1 below, as Driver1 takes the passenger to the KFC, i.e., the restaurant destination, he/she may find it very difficult to drop off the passenger because of the incorrect POI suggested by POI table. Driver1 can return this feedback to Grab transportation team, then that POI gets corrected and updated. Sometime later, Driver2 needs to pick up a food order from the same KFC, and he/she might experience the same difficulty to quickly locate the restaurant because of the inaccurate POI returned from restaurant table. This unhappy user experience could be avoided if the same restaurant and POI entities are matched up so that any update or change for an entity from one table can be automatically transferred to another.

Another use case that can benefit from matching restaurant and POI is the deduplicated search of same place entity inside both tables. A user searching for a specific restaurant will trigger the search action in both the restaurant and the POI tables. If this restaurant has a matched POI, and this matching relation has been identified, one of the two searches could be avoided, allowing a faster return of the search results and avoiding duplication of results for users.

Motivated by the above use cases, we attempted to conduct entity matching between restaurant and POI tables. The whole process includes several subtasks such as data exploration, preprocessing, distance metrics calculation, labelling, as well as supervised learning. These will be elaborated in the following sections.

## 2 PRIOR WORK IN RECORD LINKAGE FOR SPATIAL DATA

The problem of determining whether two records refer to the same or different entities is called Record Linkage or Entity Resolution, and sometimes Entity Conflation. As a literary problem it goes back for several centuries, with questions such as the identity of the poet Homer ("Were the Iliad and the Odyssey written by the same person?"). The formal study of entity resolution goes back to the 1940s, leading to a canonical statistical formulation for aligning medical records being attributed to (Newcombe et al., 1959) (see also (Talburt, 2011) for an in-depth summary).

### 2.1 Entity Resolution in Metric Spaces

A typical entity resolution approach is to model each entity using a collection of features, to train a similarity or distance function based on these features, and then to merge entities whose similarity is above some threshold, perhaps transitively using a clustering algorithm. That is, given a distance function between records, a simple application of this to entity resolution comes from assuming that two records represent the same entity if the distance between them is beneath some threshold. (Conversely, for a similarity function, if the similarity between the two, is above some threshold.)

This process is sometimes more reliable if the distance function satisfies the metric properties. Mathematically speaking, a metric on a space $M$ is a function $d : M \times M \to \mathbb{R}_{\geq 0}$ which satisfies the properties of symmetry ($d(x,y) = d(y,x)$), identity ($d(x,y) = 0 \iff x = y$), and the triangle inequality ($d(x,z) \leq d(x,y) + d(y,z)$ for all $x,y,z \in M$. Metric spaces can be particularly useful for record linkage problems because the notion of proximity can lend itself to sensible clustering properties. For an accessible introduction to metric spaces for informatics practitioners, including some well-known caveats, see (Widdows, 2004, Ch 4). For example, standard similarity functions such as cosine similarity are *not* metrics because $\cos(x,x) = 1$, and conceptual 'distances' often do not obey the symmetric rule. In high-dimensional spaces with many features, the triangle inequality also becomes a very weak condition that can allow distantly-related points to become transitively linked (Chávez et al., 2001).

The general notion of a distance function that combines several features was used throughout the experiments reported in this paper. However, since the problem addressed in this paper is bipartite matching between two separate datasets, a distance measure that supports clustering within datasets was not a requirement, and so no attempt was made to ensure that metric properties were satisfied.

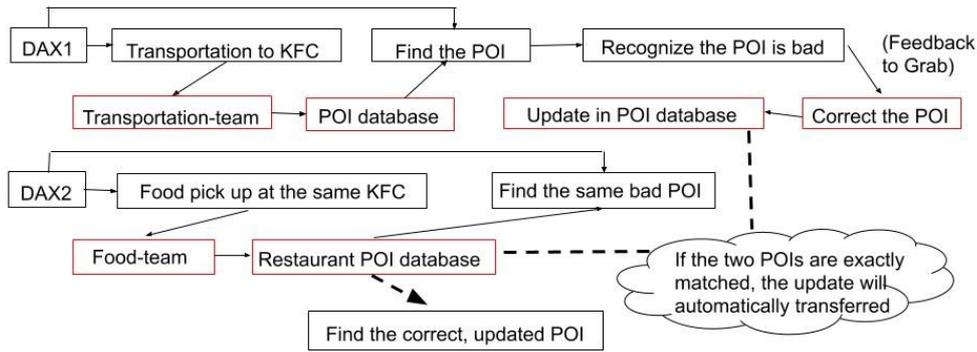

Figure 1: Use case illustration for automatic geolocation update.

Note that not all spatial record linkage problems can be posed in this similarity-based form, especially for moving objects. The challenge of recognizing the trajectories of individual moving objects from different observations arises in computational astronomy (Kubica et al., 2007), and is an increasing focus for human-generated datasets (Basık et al., 2017).

## 2.2 Blocking Features and Identifiers

When inferring pairwise matches from a distance or similarity function, comparing every record to every other record can be intractable and unnecessary. Sometimes there might be 'blocking' features, which are necessary for matching, in the sense that two records are prevented from matching if these features do not have identical values (for example, the year-of-birth of a person, if this is available in the dataset and known to be recorded accurately). Some blocking attributes may even be considered to be sufficient for matching, especially if an attribute is meant to be a unique identifier — for example, two products with the same barcode might be expected to be identical (as in (Bilenko et al., 2005)). Blocking can also be used as an early-out in computation to reduce a quadratic problem (comparing every pair of items) to a linear problem (grouping together all items with identical or at least similar values of the blocking attribute). In practice, however, it is rarely the case that any attribute can be trusted entirely with this responsibility.

## 2.3 Special Considerations for Spatial Datasets

With spatial datasets, blocking can be particularly tricky, because there is not a standard system of issuing unique identifiers. Buildings do not have ISBN numbers or Driver's License Numbers! Many have street addresses, which can sometimes be considered as a taxonomic path to finding the building (see (Widdows, 2004, Ch 3)), but these are not unique identifiers and usually have many spelling variants ("9 High Street", "9 High St", etc.). An obvious class of proposals for blocking functions for spatial datasets are sharing geographic areas — the same city, state or region, or at least the same country. However, there are two problems with relying on these as blocking functions:

- The names in these fields might not match. Surprisingly, this happens with some regularity, for example, a store near the boundary between Bellevue and Redmond may be listed in either. This problem is particularly apparent in some of our datasets for Southeast Asia.

- The input datasets might not be parsed into these fields. There are approaches for performing such parsing automatically (see e.g., (Borkar et al., 2001)). We are working on this challenge as well, though the results are not yet ready to be used as input for restaurant POI matching.

Entity resolution for spatial datasets therefore typically involves largely statistical and continuous measures of similarity. The work described here is typical in this respect, and in terms of methodology is quite similar to that of (Sehgal et al., 2006). One class of similarity is spatial proximity: provided that each entity record has latitude and longitude coordinates, the great circle distance between these points can easily be calculated and used as a feature. Another class of similarity is textual similarity: whether a string of text represents a business name, a street number, a street name, or a city name, we still expect a high similarity between two different textual descriptions of the same entity.

Another common difficulty to note is that deciding whether two records refer to the same entity — and even deciding if two entities are identical at all — is less straightforward than one might assume. Some of the problems in geo-ontology engineering, and especially problems that arise when trying to combine this with statistical methods, are discussed in (Janowicz, 2012). A restaurant and the building it occupies are conceptually two different things, as demonstrated by cases where the building takes on a different tenant, or the restaurant moves to different premises. Of course, we say "We are going to the restaurant", rather than "We are going to the building that houses the restaurant", but when we try to build formal computational models that start with such everyday conveniences, difficulties soon arise, a case-in-point being the challenge of building a formal model for reasoning from crowdsourced OpenStreetMap tags (Codescu et al., 2011).

These theoretical and practical difficulties are important to keep in mind, not because we should despair of ever solving this matching problem, but because they place sensible limits on what sort of results we should expect. This motivates the right question from a human-centered technology point of view — we are not asking "Can we build a perfect matching system", but rather "Can we build a matching system whose results will improve the experience of our users?"

## 3 DATA SOURCES USED

The two main tables this work focuses on are points-of-interest (POI) and restaurants.

### 3.1 Points of Interest

Grab uses POIs as the drop-off / pick-up location recommendations to drivers for both the transportation and food delivery purposes. The POIs are licensed externally from Google, Foursquare, Nokia, Azure, OpenStreetMap, and Aapico, and also internally by operation teams. Right now most POIs come from external sources, and Grab's own data collection activities are growing. Specific to the database of POI, each POI entity is identified by its unique id, and has dozens of attributes describing its geolocation, classification, and lifecycle milestones. The POI dataset contains approximately 150M records at the time of writing.

Efforts to deduplicate these records are ongoing, a problem that effectively subsumes the work on restaurant matching (because the restaurant dataset could be considered as just another data source).

### 3.2 Restaurants

The restaurant table contains the most updated information about those restaurants that have registered to GrabFood. Restaurant information provided by merchant owners are merged with the spatial information from Map Operation team, and then entered to database and cloud data warehouse through a series of internal platforms. Like POI, each restaurant entity has its unique id, the geolocation information, and the lifecycle milestones, but with some extra attributes such as status information and owner information, etc. The restaurant dataset contains approximately 200K records at the time of writing.

The two tables exist to support different operations, i.e., the POIs are used for transportation and the restaurants are used for food delivery.

### 3.3 Name Variation

Many of the records within these datasets vary considerably in representation. To give examples at the city and regional level, Table 1 shows some spelling differences that regularly appear with some of the larger cities in Indonesia.

Table 1: Examples of name variations found in Indonesia.

| First Name Variant | Second Name Variant | Notes |
|---|---|---|
| Jakarta | Djakarta | The spelling 'Djakarta' is usually considered obsolete but still appears |
| Solo | Surakarta | Alternative names for the same city |
| Lampung | Bandar Lampung | 'bandar' means 'city' in Malay, these are like 'New York' and 'New York City' |
| Aceh | Banda Aceh | 'banda' and 'bandar' are used similarly |
| Lubuklinggau | Lubuk Linggau | The space is optional |
| Palangkaraya | Palangka Raya | The space is optional |

## 4 MATCHING APPROACH

At a high level, our approach to matching is simple: annotate suitable (POI, restaurant) pairs as either matching or non-matching, and use this to train and evaluate classifiers that rely on basic textual and spatial features. It is important to note that the classifiers trained in this fashion classify *pairs* of (POI, restaurant) to say whether they match or not: it is not just a classifier that tells if a restaurant can be matched to any POI, but a classifier that tells us if a particular match is a good one.

This turns the matching problem into a classification problem over the Cartesian product of two sets. In practice, however, the number of potential (POI, restaurant) pairs is prohibitively large, and the chance that any randomly selected pair is a match is correspondingly small, so we sample down to make these numbers tractable and sensible by reducing the matching candidates. The steps are outlined as follows.

### 4.1 Early-Out Blocking on Geohash

Given approximately 150M POIs and 200K restaurants, there are approximately 30 trillion potential (POI, restaurant) pairs — far too many to consider all of them! To reduce the number of potential matches, we start with a blocking strategy based on geohash. A geohash is a rectangle in the latitude / longitude coordinate space, where rectangles at different levels have predictable alphanumeric identifiers (Liu et al., 2014). For these experiments, we found that taking geohashes at level 6 was a suitable tradeoff between computational performance and thoroughness. A 6-level geohash at the equator encloses an area of approximately 1.2 km × 0.6 km.

The restaurant and POI are paired up by joining on the same geohash that is of level 6. A holistic comparison between each restaurant to all the other POIs is unnecessary because two locations that are thousands of miles away could never be the identical entities. We adopted geohash to join a restaurant only to POIs that have the same geohash, thus reducing the comparison space dramatically.

### 4.2 Features Used

The spatial and non-spatial attributes used to identify the same place entities include latitude and longitude, place name, and street name, which are common fields in the POI and restaurant tables. Some distance metrics were derived from the above attributes. The distance metrics are great circle distance calculated from latitude and longitude, Levenshtein and Jaro distances for place name, and Levenshtein distance for street name. These metrics are the input features for machine learning model, and a restaurant-POI pair with lower distance values is more likely to be identical place entity. Levenshtein distance is calculated by counting the number of operations needed to convert one string into another and the edit operations include adding, deleting, and replacing. The Jaro similarity is another string similarity measure that has been used successfully for name-matching (see (Cohen et al., 2003)). It is defined as follows:

$$Jaro(\theta_1, \theta_2) = \frac{1}{3}\left(\frac{c}{|\theta_1|} + \frac{c}{|\theta_2|} + \frac{c - \frac{t}{2}}{c}\right) \quad (1)$$

where θ$_i$ are the strings, $c$ is the number of characters that match within a given distance, and $t$ is the number of transpositions needed to put the overlapping characters back in the same order. Jaro distance is derived from Jaro similarity by subtracting Jaro similarity from the value of 1. Both Levenshtein and Jaro distances are character level based, but Jaro distance focuses more on local similarities between two strings. For example, Levenshtein distance between 'ab' and 'ba', which is normalized by the sum of their lengths, is 0.5, and Jaro distance between 'ab' and 'ba' is 1 because within half of the string length we cannot find any identical character pairs from the two strings.

Some preparations are conducted for name and street address strings to convert characters to lower case, remove blank spaces and special characters, etc. Notably, the Levenshtein distance between two strings is normalized by the sum of lengths of two strings. This is because Levenshtein distance as a metric is more generous to shorter strings. For instance, Levenshtein distance between 'a' and 'b' is 1, which is seemingly smaller than the Levenshtein distance (i.e., 3) between 'abczzzzzzzzzzzzzzz' and 'fghzzzzzzzzzzzzzz'. However, we tend to recognize that the later pair is more similar compared to the similarity for 'a' and 'b'.

Great circle distance and Levenshtein distance for street name are more associated with spatial closeness of a restaurant and a POI. Nevertheless, being spatially close does not necessarily mean two place entities are identical. This is particularly the case where stores usually get crowded in densely populated metropolis such as Singapore. Therefore, similarity between names for two places is an important feature to be incorporated into our evaluation system. We consider both Levenshtein and Jaro distances for place name to increase the weight of non-spatial attribute in determining place entity similarity.

Figure 2 displays distributions of features for the different countries (i.e., Indonesia, Malaysia, Singapore, and Philippines). The samples plotted in the figure is a subset randomly selected from the final population of POI-restaurant pairs. The final population was obtained after applying early-out blocking on geohash and a few other predefined rules that help to reduce the unmatched POI-restaurant population size (which will be elaborated in the section 4.4). Due to the predefined downsampling rules there is remarkable clustering of geolocation distance within 200 meters and the clear cutoff of Levenshtein distance for name at 0.4 (i.e., the threshold we used to filter out POI-restaurant pair samples that possess low probability to be identical place entities). Interestingly, while the distribution of Levenshtein distance for name tends to be skewed to the high-value end, Jaro distance for name mimics more a bell shaped distribution, which indicates that for this specific study when the same group of string pairs are evaluated using Jaro distance instead of Levenshtein distance, a greater portion of POI name and restaurant name are considered to be more similar. This is the true case for two strings like 'ab' and 'abcd', for which the Levenshtein distance is 1/3 whereas the Jaro distance is 1/6. We observed significant number of POI name and restaurant name pairs falling into this format category, i.e., in the case where a POI refers to the same place as a restaurant the POI name is an exact substring of the restaurant name because a restaurant name is usually appended with a street address that is not included in the POI name.

### 4.3 Countries Considered

The countries where Grab offers both transportation and food delivery services that were considered for this study are Indonesia, Malaysia, Singapore, Philippines, Vietnam and Thailand. The model training and results reported here are restricted to the first four of these. Vietnam and Thailand were not investigated in this pilot study due to the difficulty of the character sets. See section 6 for further discussion of this point.

### 4.4 Annotation / Labeling with Downsampling

This project aims to match restaurant and POI through a supervised learning process. Since labels are not available for the restaurant-POI pairs joined on the same geohash, i.e, we do not know in prior whether a restaurant and a POI is matched or not, manual annotation is a necessary step before training/testing the models. We decide the relationship between a restaurant and a POI mainly based on latitude&longitude (i.e., distance), name, street.

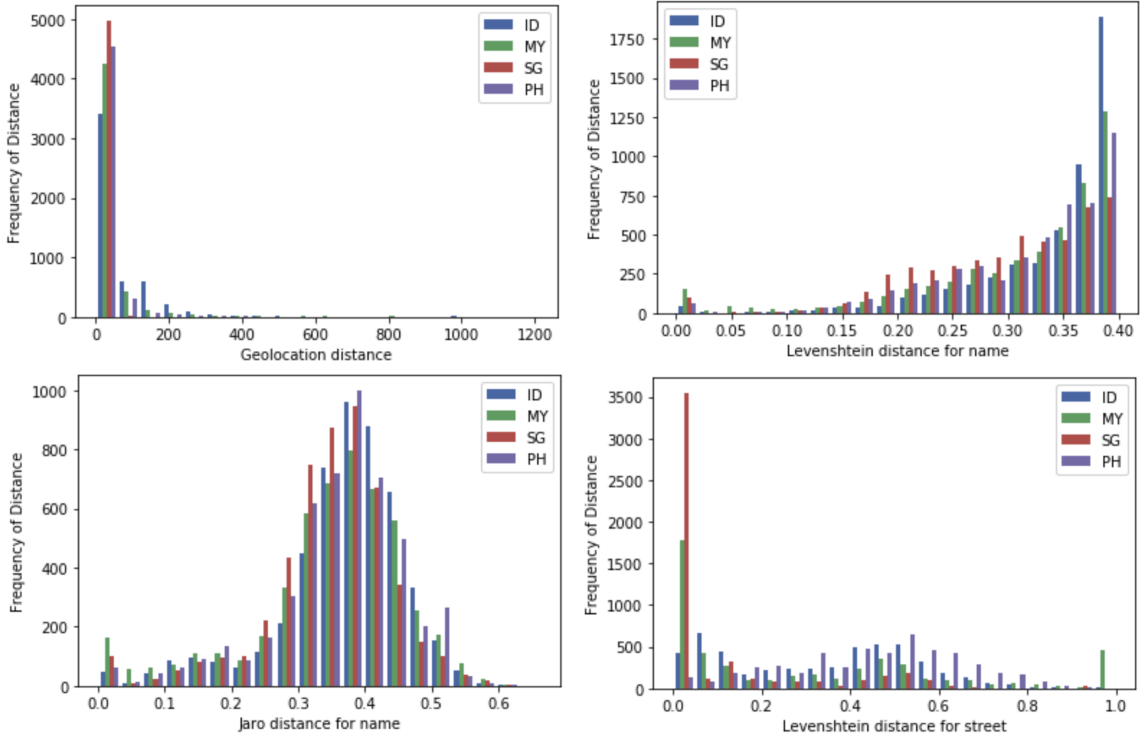

Figure 2: Histograms for features of geolocation distance (in meters), Levenshtein distance, and Jaro distance (ID: Indonesia; MY: Malaysia; SG: Singapore; PH: Philippines).

The relationship between a POI and a restaurant can be classified into two categories (see Table 2). Two place entities should refer to the same one if they have similar name and similar street address or spatially close locations. On the other hand, if it is obvious that the restaurant and the POI cannot be the same place given the different names, even though they are spatially close, they are marked as not matching.

Even after applying geohash to block out the unmatched restaurant-POI pairs, the percentage of matched restaurant-POI pairs out of the total pairs is still expected to be very low. Take Singapore for example, the total population of restaurant-POI pairs for comparison is about 220 millions and total number of unique restaurant entity is about 8K. Suppose every restaurant can find a matched POI, then the matching percentage is still as low as 0.0037%. Therefore, the whole restaurant-POI pairs population turns out rather imbalanced with the much lower rate of occurrence of matched samples. This poses the difficulty in the manual annotation, as one has to label more than 10000 pairs of restaurant-POI to hopefully get about 37 matched samples. Manually labeling 10000 pairs of samples requires too much human labor, which is both time consuming and error prone.

We proposed two predefined rules to exclude or downsample a great portion of restaurant-POI pairs that are expected to be unmatched. This is based on two underlying assumptions. The first one is that a POI that is closer to a restaurant is more likely to be matched with the restaurant than another POI that is further away. Therefore, we only consider the top K nearest POIs as the potential matching candidates. Secondly, a restaurant-POI pair with name Levenshtein distance greater than 0.4 tend to be unmatched. A special case is 'abc' and 'def', which has the normalized Levenshtein distance of 0.5. As one of the strings gets longer, that ratio will be higher than 0.5, moving further from being matched. We chose 0.4 as the threshold to apply stricter penalty to name edit distance. After prescreening the restaurant-POI pairs based on the above predefined rules, the whole population for comparison is reduced significantly, i.e., about 32K. This is beneficial to identifying considerable number of matched restaurant-POI pairs without having to label a huge amount of pairs.

Table 2: Restaurant-POI relationship types.

| Matched | Name | Street | Distance (meters) |
|---|---|---|---|
| Restaurant | fore coffee - bintarof | jl. boulevard bintaro jaya ruko kebayoran arcade 2 blok b3 no 51 pd. jaya pd. aren tangerang selatan | 0.0 |
| POI | fore coffee - 20fit bintaro | jl. boulevard bintaro jaya ruko kebayoran arcade 2 blok b3 no 51 pondok aren | |
| Unmatched | Name | Street | Distance (meters) |
| Restaurant | mie setan 'noodle and dimsum' - tlogomas | jl. raya tlogomas no. 31 tlogomas lowokwaru malang | 15.92 |
| POI | KFC- tlogomas | jalan raya tlogo mas | |

The manual labeling follows three steps. We first randomly select 500 samples from the population for comparison, compare through each pair of restaurant and POI, and assign a label to the pair which could be matched and unmatched that is coded as 1 and 0 respectively. The 500 pre-labeled samples are applied to train and test a Decision Tree, which is used to subsequently predict for another randomly selected 2000 samples. Among the 2000 samples, the restaurant-POI pairs predicted as matched are manually rectified if any of the predicted labels are wrong. In this way, for each country we could collected over 1000 pre-labeled samples that could be used for training and testing.

### 4.5 Model Training

Each dataset was divided between training and testing with the ratio of 4:1, that is, 20% of the datasets were held aside for evaluation. All models were trained using Scikit Learn in Python (Pedregosa et al., 2011). The implementation uses PySpark for distributed computing where appropriate.

### 4.6 Computational Resources

Preprocessing to get the final population of restaurant-POI pairs appeared to be the most computationally time consuming and memory demanding process. This includes but not limited to joining restaurants and POIs based on geohash, calculating distance metrics, applying predefined rule-out conditions to decrease unmatched pair population, saving parquet and CSV files to AWS S3. These operations were conducted on a Spark cluster of 20 machines. Although suitable partition strategies were utilized, depending on the population sizes of restaurant and POI for the different countries, running time on Spark could last for a few hours, with the longest to be 4 hours for Indonesia. 10 GB and 40 GB memory was allocated to master and executors respectively to evade the out-of-memory failure. Since the final comparison domain was considerably reduced, training and testing the tree-based models and making predictions ran very quickly and lasted for several minutes.

## 5 RESULTS AND DISCUSSION

This section reports the results for each country for which supervised classifiers were trained and tested (as shown in Table 3). Here Class 1 refers to the unmatched and Class 2 to the matched pairs. As can be seen, results of test accuracy are uniformly quite high regardless of countries. The lowest accuracy is still over 93%, a great improvement from a random guess based on the proportion of major category (i.e., non-matching class taking up 70-80% of the population). A comparison of precision and recall across the different countries shows that Malaysia has the weakest non-matching results, and the weakest matching results are found in the Philippines. The tree-based classifiers perform the best for matching restaurants and POIs in Indonesia in terms of precision and recall scores for the matched class. This is demonstrated in Figure 3 that lists some randomly selected samples of restaurant-POI pairs predicted as matched in Indonesia. After examining on Google map, it is observed that all the pairs are correctly predicted even though there are many variations in names and street addresses between restaurants and POIs. Some of the cases are hazy to be identified as matched when checking with human eyes. For example, the first pair has the quite different street address expressions and the geolocation distance is also relatively further than the other pairs. However, the model could still make a correct prediction through the experience it learned from the training set. Some POIs have void street addresses, but the model is able to recognize the matched pairs based on the close geolocation distance combined with similar name strings.

Using Random Forest to predict the matching of all restaurant-POI candidate pairs, we estimated the number of matchable restaurants in each country considered. These were approximately 37% (Indonesia), 53% (Malaysia), 51% (Philippines) and 23% (Singapore). On average 38% of restaurants in the four countries can be found to match with at least one POI, implying a great potential to enhance user experience for drivers and passengers if those restaurants happen to be the popular ones. The reason for the low matching rate in Singapore is still under investigation. It is important to note that the precision and recall scores measure the recall of the automatic classifier compared with the manual matching efforts during the annotation process. So for example, achieving high 90's recall for Malaysia means that of the whole population of matched restaurant-POI pairs, over 90% were also captured by the classifier. It does not mean that over 90% of restaurants have been matched to a corresponding POI.

Notably, the role played by each input feature weights differently when Random Forest was used to classify the relationship between merchants and POIs,
as shown in 4. For each country, the same order of the importance of features can be observed, i.e. Jaro distance of name appears the most important, followed by Levenshtein distance of name, then great circle geolocation distance, and minimal role played by Levenshtein distance of street name. It is an interesting finding that Jaro distance of name outweighs Levenshtein distance of name in determining if a merchant-POI pair is matched or not. Compared to Levenshtein distance, Jaro distance focuses more on local similarity of two strings and is more likely to assign lower distance score to two strings with identical substrings. As a result of the fact that many restaurant entries have their name concatenated with a street name (as discussed in the section 4.2, Jaro distance turns out to be a better measurement of the similarity between restaurant and POI names). As for geolocation distance, its much smaller importance score seems surprising. However, remember that some previous early-out blocking actions were already executed on the geolocation distance, so the real role of geolocation distance should turn out to be much more prominent than what the figure shows.

A map representing GrabFood restaurants in Singapore is displayed in Figure 5. The red dots stand for restaurants that do not have matched POIs, and the green dots represent restaurants that have matched POIs. Spatial distribution of the two categories of restaurants shows that there is no special clusters of certain type of restaurant, and the restaurants with/without matched POIs are mixed with each other.

Finally, we ran an experiment to see if results improved or deteriorated when datasets from each country were not kept separate. We expected that the extra data for training could be beneficial, but that using training data from another country could lead to poorer results from inappropriate extrapolation. What we found is that on the whole results became less reliable (Table 3), with the exception being that recall rate of matching results in Malaysia improved a little. One hypothesis is that the similarity between Malaysian and Indonesian addresses and street names may contribute to the usefulness of Indonesian data for Malaysia. The impaired matching result after merging all countries together could be attributed to the different distributions of features across the different countries (see Figure 2). For instance, Levenshtein distance between names for Singapore is distributed relatively more homogeneously, whereas for other countries there is concentration of high value Levenshtein distance of name. Distributions of Jaro distance for name do not exactly coincide with each other for the different countries, so do for the distributions of Levenshtein distance for street name. Hence, there is regional variations across the different countries we are working on in terms of spatial distribution patterns of restaurant vs. POI, the name and street address format conventions, etc. This leads to the consequence that the Levenshtein distance for name with the value of 0.1 could indicate the matched POI and restaurant in Indonesia, but being unmatched in Singapore.

Table 3: Tree-based model performance for different countries.

| Classifiers | Test accuracy | F1 score (class1) | F1 score (class2) | Precision (class1) | Precision (class2) | Recall (class1) | Recall (class2) |
|---|---|---|---|---|---|---|---|
| Malaysia | | | | | | | |
| RandomForest | 93.1% | 95.2% | 87.6% | 96.0% | 85.7% | 94.4% | 89.6% |
| AdaBoost | 95.5% | 96.9% | 91.8% | 97.2% | 91.2% | 96.6% | 92.5% |
| GradientBoost | 94.7% | 96.4% | 90.2% | 96.1% | 90.9% | 96.6% | 89.6% |
| XGBoost | 94.7% | 96.3% | 90.5% | 97.1% | 88.6% | 95.5% | 92.5% |
| Indonesia | | | | | | | |
| RandomForest | 97.0% | 98.1% | 92.6% | 97.6% | 94.3% | 98.6% | 90.9% |
| AdaBoost | 96.6% | 97.8% | 91.9% | 98.1% | 91.1% | 97.6% | 92.7% |
| GradientBoost | 98.1% | 98.8% | 95.5% | 99.0% | 94.6% | 98.6% | 96.4% |
| XGBoost | 97.7% | 98.6% | 94.5% | 98.6% | 94.5% | 98.6% | 94.5% |
| Philippines | | | | | | | |
| RandomForest | 97.0% | 98.2% | 90.1% | 99.6% | 83.7% | 96.9% | 97.6% |
| AdaBoost | 96.3% | 97.8% | 88.2% | 99.6% | 80.4% | 96.1% | 97.6% |
| GradientBoost | 96.7% | 98.0% | 89.1% | 99.6% | 82.0% | 96.5% | 97.6% |
| XGBoost | 97.0% | 98.2% | 90.1% | 99.6% | 83.7% | 96.9% | 97.6% |
| Singapore | | | | | | | |
| RandomForest | 97.6% | 98.6% | 90.6% | 98.3% | 92.3% | 98.9% | 88.9% |
| AdaBoost | 96.2% | 97.8% | 85.7% | 98.3% | 82.8% | 97.2% | 88.9% |
| GradientBoost | 97.1% | 98.3% | 89.3% | 98.9% | 86.2% | 97.8% | 92.6% |
| XGBoost | 97.6% | 98.6% | 90.9% | 98.9% | 89.3% | 98.3% | 92.6% |
| Merge all four countries | | | | | | | |
| RandomForest | 95.9% | 97.5% | 87.2% | 98.4% | 83.1% | 96.6% | 91.7% |
| AdaBoost | 95.4% | 97.2% | 86.0% | 98.6% | 80.4% | 95.9% | 92.3% |
| GradientBoost | 95.5% | 97.3% | 86.3% | 98.7% | 80.6% | 95.9% | 92.9% |
| XGBoost | 95.6% | 97.3% | 86.6% | 98.7% | 81.0% | 96.0% | 92.9% |

# 6 CHALLENGES WITH THAI

Matching restaurants and POIs in Thailand is more challenging than in the countries evaluated above, because of the non-Roman character set. Other major languages in countries where Grab operates that use non-Roman character sets include Burmese and Khmer, while Vietnamese script is based on Roman characters with many diacritical accent and tone marks. Each of these scripts is challenging for unfamiliar readers and computers to process. This section considers issues with Thai which are somewhat repre-sentative of the kinds of problems encountered across different languages and especially different scripts.

| id | poi_id | name | poi_name | street | poi_street | distance |
|---|---|---|---|---|---|---|
| 6-CYJ2RLC1E7ADLA | IT.2LVYTPDW55AM9 | warung hana - sukawati | warung hana | jl. goa gajah raya sukawati gianyar | jl. teges kanginan peliatan ubud | 31.043903 |
| 6-CYK2GFDTG3AFG6 | GA.33URHY68527FD | mie setan 'noodle and dimsum' - tlogomas | mie setan tlogomas ok | jl. raya tlogomas no. 31 tlogomas lowokwaru ma... | jalan raya tlogomas | 10.719604 |
| 6-CYK3JYUDUGKERE | GA.2ZTXZRD90XDTV | dum dum thai drinks - bandung electronic center | dum dum thai drinks bec | bandung electronic center lt. 3a jl. purnawarm... | NaN | 3.321889 |
| 6-CYKALPKHJY4XE2 | IT.24DWR72PW4PE9 | kebab mama putri - dukuh | pecel lele mama putri dukuh | jl.dukuh v no. 10 gang f dukuh kramatjati jak... | jl. dukuh v no.10 dukuh kramatjati | 12.298162 |
| 6-CYKGLYVJCCLUL2 | IT.0SELTYEH85FJW | fore coffee - bintaro | fore coffee - 20fit bintaro | jl. boulevard bintaro jaya ruko kebayoran arca... | jl. boulevard bintaro jaya ruko kebayoran arca... | 0.000000 |
| 6-CYKGLYVJCCLUL2 | GA.3AW9Z8MDIJMVT | fore coffee - bintaro | fore coffee - 20fit bintaro | jl. boulevard bintaro jaya ruko kebayoran arca... | jalan boulevard bintaro jaya | 17.419987 |
| 6-CYKUPFWCE3CXTA | GO.19EX9ED03CZPC | kedai seblak rajanya pedas - sarijadi | kedai seblak rajanya pedas | sariwangi jl. sarijadi raya ( depan flat blok ... | jalan sarijadi raya | 13.484235 |
| 6-CYKXTU3XN7B3A2 | NK.0P1K2GQG8HAM4 | rm uda juo - bencongan | rm uda juo | perumahan harapan kita jl. alamanda vii blok j... | NaN | 11.265291 |
| 6-CYL2GPWBKE4VKA | IT.1Q3ZMMB77Q3X5 | bengkel camilan - mesuji 1 | bengkel camilan mesuji 1 | jl. mesuji 1 c2 no 3385 demang lebar daun ilir... | jl. mesuji 1 c2 no 3385 demang lebar daun ilir... | 13.466086 |
| 6-CYL2GPWBKE4VKA | GA.3UXT22S0U8OAU | bengkel camilan - mesuji 1 | bengkel camilan | jl. mesuji 1 c2 no 3385 demang lebar daun ilir... | NaN | 14.669184 |

Figure 3: Example of restaurant-POI pairs predicted as matched for Indonesia.

## 6.1 Transliteration vs. Translation

Non-Roman character sets are more challenging not just because it makes the characters harder for unfamiliar users, but because it influences the structure of names, which often contain repetitions of local and Roman character versions. For example, the name 'Starbucks' may appear in a title repeated in Roman / English and Thai characters, as in 'Starbucks (สตาร์บัคส์)', which makes string overlap techniques behave differently. This also illustrates the point that translating business names and addresses as if they were regular paragraph texts often works badly — for example, a good translation of 'Starbucks (สตาร์บัคส์)' into English would not be 'Starbucks (Starbucks)'. Proper names often should not be translated at all. For example, the (short) Thai name for Bangkok is 'กรุงเทพ' (Krung Thep), which means 'city of angels', but showing this translation to English speakers would possibly lead to confusion between Bangkok and Los Angeles! In practice, restaurants with the Thai name 'กรุงเทพ' (which are relatively common worldwide) are for more likely to be rendered as 'Krung Thep' for English speakers, which is a phonetic transliteration rather than a semantic translation.

A more familiar example for Europeans is perhaps the famous *Avenue des Champs-Élysées* in Paris. To render this as 'Elysian Fields Avenue' on an English-language streetmap would be much more likely to confuse than to inform, partly because it's less familiar, and partly because it doesn't correspond to the physical street signs.

Understanding which phrases should be translated semantically, which should be transliterated between character sets, and which should be left alone appears to be an open area for research. It is at least fair to note that current machine translation systems are not designed to recognize and respond appropriately to these differences, and should not be trusted for these purposes without manual review. For reasons such as this, adapting our methods to Vietnamese and Thai (and eventually Burmese and Khmer) is left for future work.

## 6.2 Non-Symmetric Overlap Measures

The relationship between the name 'Starbucks' and 'Starbucks (สตาร์บัคส์)' is clearly an asymmetric containment relationship. Symmetric measures such as Levenshtein and Jaro do not naturally capture this (though as remarked above, Jaro gives more credit for matching substrings). In other work at Grab for machine translation and intent recognition for search, it is also apparent that the relationship between a source and a target (for example, a food query and a menu item) is often asymmetric. Symmetric distance measures sometimes can be enhanced by taking this into account — for example, smartphone users in informal situations often leave out diacritical marks in Vietnamese and vowels in Indonesian. A thorough analysis of such phenomena is beyond the scope of this paper.

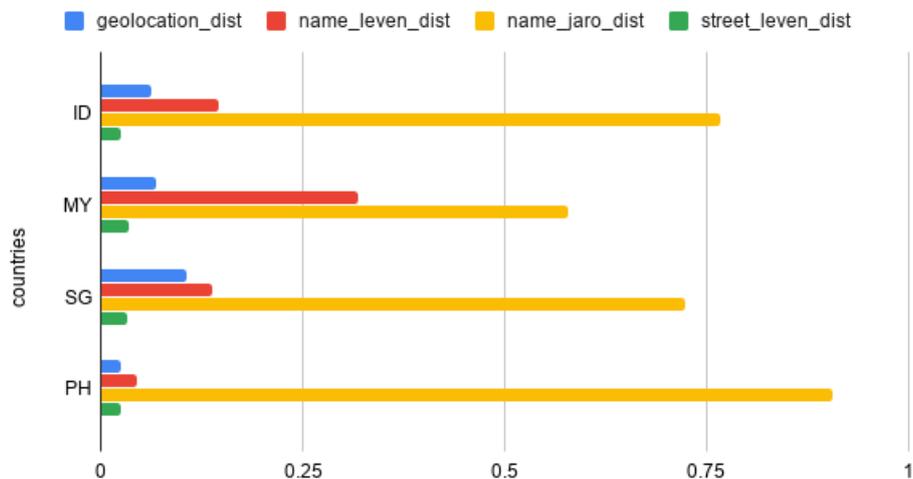

Figure 4: Feature importance scores in classification using Random Forest.

# 7 CONCLUSIONS

This paper has demonstrated that relatively good matching results between restaurants and points of interest in Southeast Asia can be obtained by training tree-based classifiers iteratively on a few hundred pairs. This is an encouraging start. These results approximate human performance at finding matching pairs given available search tools. An open question this leaves is how close this gets to exhausting the matching possibilities — when no match for a restaurant is found in the POI dataset, does this indicate that matching need to be improved, or that the POI data itself is lacking? If the latter is the cause of no-match restaurants, a next step may be just to add the unmatched restaurants as POIs.

Many local questions can be asked about the efficiency of raw string matching, because many address names vary systematically. For example, the word 'Jalan' in Malay languages (including Indonesian and Malaysian) means 'Street' and can be abbreviated as 'Jl' and 'Jln' without any change in meaning. Such variants should be considered identical for matching. This is also an example of a general question for knowledge discovery and natural language processing in Southeast Asia — when is it helpful to treat Indonesian and Malaysian separately, and when is it useful to treat them as part of the same language group?

Finally, we have clearly worked only with the Southeast Asian countries and languages that use the Roman alphabet (without extensive diacritic modifications, as in Vietnamese). Adapting this work to other countries with different alphabets and various transliteration methods will be important for extending coverage throughout the region.

# ACKNOWLEDGEMENTS

The authors would like to thank colleagues from Grab for guidance and help throughout this work, including Xiaonan Lu, Kristin Tolle, Jagan Varadarajan, Wenjie Xu, Sien Yi Tan, Sidi Chang, and Jacob Lucas.

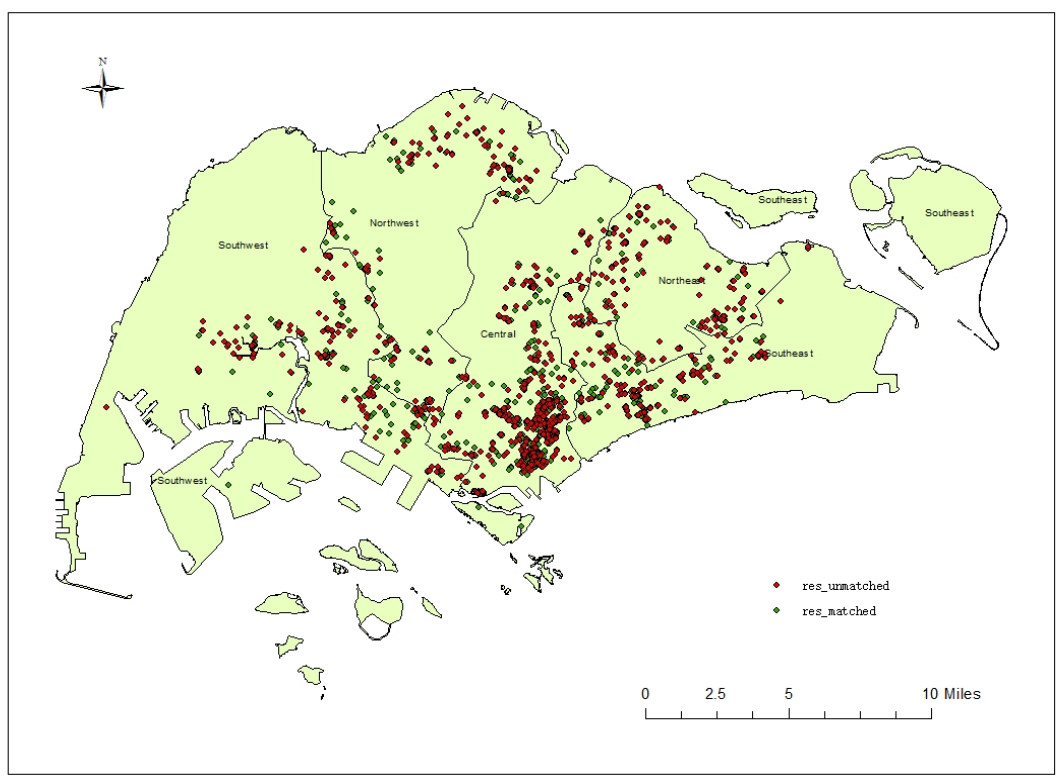

Figure 5: Spatial distribution of sampled restaurants with/without matched POIs in Singapore.

## References

Basık, F., Gedik, B., Etemoğlu, Ç., and Ferhatosmanoğlu, H. (2017). Spatio-temporal linkage over location-enhanced services. *IEEE Transactions on Mobile Computing*, 17(2):447–460.

Bilenko, M., Basil, S., and Sahami, M. (2005). Adaptive product normalization: Using online learning for record linkage in comparison shopping. In *Fifth IEEE International Conference on Data Mining (ICDM'05)*. IEEE.

Borkar, V., Deshmukh, K., and Sarawagi, S. (2001). Automatic segmentation of text into structured records. In *ACM SIGMOD Record (Vol. 30, No. 2, pp. 175-186)*. ACM.

Chávez, E., Navarro, G., Baeza-Yates, R., and Marroquín, J. L. (2001). Searching in metric spaces. *ACM computing surveys (CSUR)*, 33(3):273–321.

Codescu, M., Horsinka, G., Kutz, O., Mossakowski, T., and Rau, R. (2011). Osmonto — an ontology of OpenStreetMap tags. *State of the map Europe (SOTM-EU)*.

Cohen, W. W., Ravikumar, P., and Fienberg, S. E. (2003). A comparison of string distance metrics for name-matching tasks. In *2003 International Conference on Information Integration on the Web (pp. 73-78)*. II-Web.

Janowicz, K. (2012). Observation-driven geo-ontology engineering. *Transactions in GIS*, 16(3):351–374.

Kubica, J., Denneau, L., Grav, T., Heasley, J., Jedicke, R., Masiero, J., Milani, A., Moore, A., Tholen, D., and Wainscoat, R. J. (2007). Efficient intra-and inter-night linking of asteroid detections using kd-trees. *Icarus*, 189(1):151–168.

Liu, J., Li, H., Gao, Y., Yu, H., and Jiang, D. (2014). A geohash-based index for spatial data management in distributed memory. In *22nd International Conference on Geoinformatics (pp. 1-4)*. IEEE.

Newcombe, H. B., Kennedy, J. M., Axford, S. J., and James, A. P. (1959). Automatic linkage of vital records. *Science*, 130(3381):954–959.

Pedregosa, F., Varoquaux, G., Gramfort, A., Michel, V., Thirion, B., Grisel, O., Blondel, M., Prettenhofer, P., Weiss, R., Dubourg, V., and Vanderplas, J. (2011). Scikit-learn: Machine learning in python. *Journal of machine learning research*, 12:2825–2830.

Sehgal, V., Getoor, L., and Viechnicki, P. D. (2006). Entity resolution in geospatial data integration. In *14th annual ACM international symposium on Advances in geographic information systems (pp. 83-90)*. ACM.

Talburt, J. R. (2011). *Entity resolution and information quality*. Morgan Kaufmann, San Francisco.

Widdows, D. (2004). *Geometry and meaning (Vol. 773)*. Stanford: CSLI publications.